\documentclass[letterpaper]{article} 
\usepackage{aaai24}  
\usepackage{times}  
\usepackage{helvet}  
\usepackage{courier}  
\usepackage[hyphens]{url}  
\usepackage{amsmath}
\usepackage{graphicx} 
\usepackage{natbib}  
\usepackage{caption} 
\usepackage{url}            
\usepackage{booktabs}       
\usepackage{amsfonts}       
\usepackage{nicefrac}       
\usepackage{microtype}      
\usepackage{xcolor}         
\usepackage{svg}
\usepackage{colortbl}

\usepackage[export]{adjustbox} 

 \usepackage{subcaption}
    \usepackage{listings}
\definecolor{keywords}{rgb}{0,0,0.7}
\definecolor{maroon}{cmyk}{0, 0.87, 0.68, 0.32}
\definecolor{halfgray}{gray}{0.55}
\definecolor{ipython_frame}{RGB}{207, 207, 207}
\definecolor{ipython_bg}{RGB}{247, 247, 247}
\definecolor{ipython_red}{RGB}{186, 33, 33}
\definecolor{ipython_green}{RGB}{0, 128, 0}
\definecolor{ipython_cyan}{RGB}{64, 128, 128}
\definecolor{ipython_purple}{RGB}{170, 34, 255}
\definecolor{listing-keyword}{HTML}{435489}
\definecolor{listing-keyword-2}{HTML}{1284CA}
\definecolor{listing-keyword-3}{HTML}{914661}
\lstdefinestyle{mystyle}{
    basicstyle=\ttfamily,                    
    showspaces=false,                
    showstringspaces=false,
    showtabs=false,                  
    tabsize=2,
    keywordstyle={\color{listing-keyword}},
    keywordstyle={[2]\color{listing-keyword-2}},
    moredelim = [s][\color{listing-keyword}]{train\ }{input},
    moredelim = [s][\color{listing-keyword-3}]{test\ }{input},
    moredelim = [s][\color{listing-keyword-3}]{Repeat\ on\ all\ test\ }{points},
    literate={0}{{{\color{listing-keyword-2}0}}}1
             {1}{{{\color{listing-keyword-2}1}}}1
             {2}{{{\color{listing-keyword-2}2}}}1
             {3}{{{\color{listing-keyword-2}3}}}1
             {4}{{{\color{listing-keyword-2}4}}}1
             {5}{{{\color{listing-keyword-2}5}}}1
             {6}{{{\color{listing-keyword-2}6}}}1
             {7}{{{\color{listing-keyword-2}7}}}1
             {8}{{{\color{listing-keyword-2}8}}}1
             {9}{{{\color{listing-keyword-2}9}}}1
             {\#XX}{{{\color{listing-keyword-2}\#XX}}}3
             {False}{{{\color{maroon}False}}}5
             {True}{{{\color{ipython_green}True}}}4
}

\def\b0{\mathbf{0}}
\def\b1{\mathbf{1}}

 \newcommand{\black}[1]{{\color{black}{#1}}}

\usepackage{graphicx}
\usepackage{xcolor}
\usepackage{booktabs}
\usepackage{amsthm, amsmath, amssymb}
\usepackage{mathtools}
\usepackage{graphicx,subcaption,float}
\usepackage{breqn}
\usepackage[autostyle]{csquotes}
\usepackage[british]{babel}
\usepackage{breakcites}
\usepackage{multirow}
\usepackage{multirow}
\usepackage{comment}

\usepackage{algorithm}
\usepackage{algorithmic}

\usepackage{newfloat}
\usepackage{listings}
\DeclareCaptionStyle{ruled}{labelfont=normalfont,labelsep=colon,strut=off} 
\floatstyle{ruled}
\newfloat{listing}{tb}{lst}{}
\floatname{listing}{Listing}
\title{How Robust are LLMs to In-Context Majority Label Bias?}
\author{
    Karan Gupta\equalcontrib, 
    Sumegh Roychowdhury\equalcontrib,
    Siva Rajesh Kasa\equalcontrib, \\
    Santhosh Kumar Kasa,
    Anish Bhanushali,
    Nikhil Pattisapu,
    Prasanna Srinivasa Murthy
}
\affiliations{
    \textsuperscript{\rm} Amazon\\

    \{karaniis,sumegr,kasasiva\}@amazon.com
}

\usepackage{bibentry}

\begin{document}

\maketitle

\begin{abstract}
 In the In-Context Learning (ICL) setup, various forms of label biases can manifest. One such manifestation is \textit{majority label bias}, which arises when the distribution of labeled examples in the in-context samples is skewed towards one or more specific classes making Large Language Models (LLMs) more prone to predict those labels. Such discrepancies can arise from various factors, including logistical constraints, inherent biases in data collection methods, limited access to diverse data sources, etc. which are unavoidable in a real-world industry setup. In this work, we study the robustness of in-context learning in LLMs to shifts that occur due to majority label bias within the purview of text classification tasks. Prior works have shown that in-context learning with LLMs is susceptible to such biases. In our study, we go one level deeper and show that the robustness boundary varies widely for different models and tasks, with certain LLMs being highly robust ($\sim$90\%) to majority label bias. Additionally, our findings also highlight the impact of model size and the richness of instructional prompts contributing towards model robustness. We restrict our study to only publicly available open-source models to ensure transparency and reproducibility.
\end{abstract}

\section{Introduction}
\label{intro}

Large language models (LLMs) have demonstrated notable capabilities in effectively executing unseen tasks solely based on provided prompts \cite{brown2020language}. This research investigates the in-context learning (ICL) paradigm for text classification (TC) tasks, a significant area of interest within Natural Language Processing (NLP). The ICL approach in LLMs has demonstrated the capability to achieve performance similar to that of the fine-tuned approach, attributed to their size and pre-training tasks \citep{wang2023investigating, sun2023text}.

However, ICL is highly depended on the design of the in-context prompt, including the selection \cite{liu2021makes}  and order of in-context examples \cite{lu2022fantastically}. \citet{zhao2021calibrate} discussed how the instability of ICL results is often due to biases in the model's predictions. One such example is the \textit{recency bias} where the model prediction favors the label of the last in-context example. They also discuss about \textit{majority label bias} where GPT-3 \cite{brown2020language} tends to prefer answers that are frequent in the in-context prompt.
\citet{fei2023mitigating} also explored label biases in ICL and introduces three types of label biases: \textit{vanilla}, \textit{context}, and \textit{domain} biases. \textit{Vanilla label} bias refers to the model's inclination to predict specific label names without considering the context. One potential factor contributing to this bias is the label name term frequencies in the pre-training corpus. \textit{Context label} bias arises from the influence of context prompts (e.g. LLMs tend to exhibit a preference for the predominant and/or final label among the in-context examples). \textit{Domain label} bias means that the type of task the model is working on can affect its predictions. For example, if a LLM is trying to figure out if a patient is sick or healthy based on medical descriptions, it might be biased towards predicting ``sick" because words often used in those descriptions are more connected to health problems. This bias can make the model consistently favor certain predictions.
\citet{fei2023mitigating} also introduced Domain-context Calibration (DC) for LLMs to mitigate these biases by estimating label bias using random words from the unlabeled evaluation dataset and adjusting predictions accordingly, effectively addressing domain-label bias.

In this work, our emphasis is on understanding \textit{majority label bias} \cite{zhao2021calibrate}, a type of context label bias \cite{fei2023mitigating} that arises when in-context prompts contain relatively more examples from one class compared to the others.
\citet{zhao2021calibrate} showed that the performance of ICL using LLMs for TC is susceptible to extreme class imbalance in the prompt. However, they have not exhaustively examined how robust the model performance is to varying proportions of different classes in the prompt in TC tasks. In this study, we address it by studying ICL performance of various LLMs across diverse proportions of different class samples in the prompt. Interestingly, our study reveals that ICL using certain LLMs exhibit remarkable level of robustness with approximately $\sim 90\%$ resistance to majority label bias.

Our contributions are as follows: \textbf{(a)} We conduct a comprehensive study on the robustness boundaries of in-context evaluations using pre-trained LLMs to majority label bias. Our findings shed light on the model's performance under varying distributional conditions. 
\textbf{(b)} We provide ablations showing the effect of model sizes and the impact of adding informative instruction prompts on model robustness to majority label bias.

\begin{figure*}[htb]
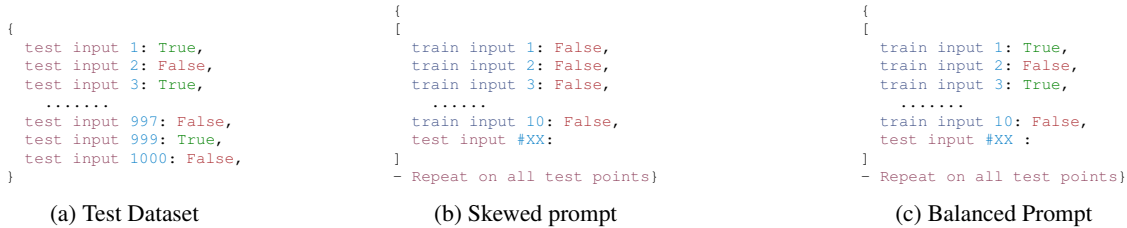

        \centering
        \begin{subfigure}[b]{.3\linewidth}
            \tiny
            \centering
            \begin{tabular}{c}
\begin{lstlisting}[style=mystyle, numbers=none]
{
  test input 1: True,
  test input 2: False,
  test input 3: True,
    .......
  test input 997: False,  
  test input 999: True,  
  test input 1000: False,
}
\end{lstlisting}
            \end{tabular}
            \caption{Test Dataset}
        \end{subfigure}%
        \begin{subfigure}[b]{.3\linewidth}
        \tiny
            \centering
            \begin{tabular}{c}

\begin{lstlisting}[style=mystyle, numbers=none]
{
[
  train input 1: False,
  train input 2: False,
  train input 3: False,
    ......
  train input 10: False,
  test input #XX: 
]
- Repeat on all test points}
\end{lstlisting}
             \end{tabular}
             \caption{Skewed prompt}
        \end{subfigure}%
        \begin{subfigure}[b]{.4\linewidth}
        \tiny
            \centering
            \begin{tabular}{c}
\begin{lstlisting}[style=mystyle, numbers=none]
{
[
  train input 1: True,
  train input 2: False,
  train input 3: True,
    .......
  train input 10: False,
  test input #XX : 
]
- Repeat on all test points}
\end{lstlisting}
            \end{tabular}
            \caption{Balanced Prompt}
        \end{subfigure}
        \caption{\small \textbf{Majority label bias in In-Context Learning:} On the left, we have an illustrative test data, with 1000 inputs, which has both \texttt{True} and \texttt{False} labels. In the middle, we have a highly \textit{skewed prompt} where all the 10 examples in the prompt are \texttt{False}. On the right, we have a more \textit{balanced prompt} which has a mix of both \texttt{True} and \texttt{False} examples. Majority Label bias is relatively high in the \textit{skewed prompt} as compared to the \textit{balanced prompt}. }
                \label{Fig_1:sample_selection_bias}
    \end{figure*}

\section{Proposed Setting}
\label{proposed}

\textbf{Datasets}: We choose the \textit{BoolQ} \cite{clark2019boolq} and \textit{RTE-1/2/3} \cite{giampiccolo2007third} datasets from \textit{lm-evaluation-harness}\footnote{\url{https://github.com/EleutherAI/lm-evaluation-harness}} which have binary labels and a real-world, multi-class dataset \textit{COVID-5G Conspiracy} \cite{micallef2020role} to synthetically create prompts with varying degree of majority label bias. 

BooleanQuestions (BoolQ) dataset is a set of 12,800 \texttt{Yes/No} question-answer pairs based on real-world knowledge. The distribution of \texttt{Yes/No} in this dataset is 62 / 38\%. We split the dataset in 80-20\% proportions into train and test respectively. We provide $N$ examples from the train set in the prompt and let the LLM `predict' the answer for a new question from the test set in an auto-regressive fashion.
Each question-answer pair is accompanied by a supporting passage, but we choose to exclude this in our study. This decision stems from our focus on comprehending the implicit biases in sample selection demonstrated by LLMs when tasked with answering questions based on their world knowledge, rather than relying on deducing the answers directly from the provided passages.
We choose $N=50$ based on the average length of tokens in the dataset and the maximum context length that is supported by the LLM. Once the predictions are generated for all the datapoints in the test set, we compute the weighted F1 using the ground-truth labels. We repeat this process by varying the proportion of labels in the $N$ examples in the prompt i.e. we vary the \texttt{Yes/No} \% of the in-context prompts provided as input to the model starting from (0\% \texttt{Yes}, 100\% \texttt{No}) till (100\% \texttt{Yes}, 0\% \texttt{No}) in equal step sizes of 10\%. For example, if the total number of in-context examples is $N=50$, then we run the model for 11 settings: \{(0 \texttt{Yes}, 50 \texttt{No}), (5 \texttt{Yes}, 45 \texttt{No}), ... , (50 \texttt{Yes}, 0 \texttt{No})\}. Our objective is to assess the model's robustness (in terms of weighted F1) to majority label distribution variations within its context. 

The PASCAL Recognizing Textual Entailment (RTE-1/2/3) dataset contains 4167 premise-hypothesis pairs and the objective is to predict \texttt{Yes/No} based on whether the premise entails the hypothesis or not. We choose 2400 datapoints from 4167 as the test set and keep remaining for the train set. The distribution of \texttt{Yes/No} in this dataset is 50/50\% . We set $N=10$ based on the average size of the input. The rest of the experimental procedure is identical to that of BoolQ dataset. 

COVID-5G Conspiracy \citep{micallef2020role} dataset has 2400 tweets classified into three categories (800 tweets per class) - \texttt{Misinformation}, \texttt{Counter-misinformation} or \texttt{Irrelevant} - regarding the conspiracy theory that COVID-19 is being caused due to 5G radiation. We choose 720 datapoints from the total 2400 as the test set and keep remaining for the train set. Here we set $N=28$ based on the average tweet length in the dataset. Since the dataset has three ground-truth labels we don't evaluate for all possibilities as that'll blow up the total number of experiments. We carefully evaluate for a few settings (see Table~\ref{tab:table_covid} in Appendix) like: (100\% \texttt{Misinformation}, 0\% \texttt{Counter-misinformation}, 0\% \texttt{Irrelevant}), (0\% \texttt{Misinformation}, 50\% \texttt{Counter-misinformation}, 50\% \texttt{Irrelevant}), (25\% \texttt{Misinformation}, 25\% \texttt{Counter-misinformation}, 50\% \texttt{Irrelevant}), etc.

The default setting in our experiments is to prepend a task-specific instruction in the prompt (more details in Appendix); these experiments are denoted with \textit{w: with instruction}. This task-specific instruction helps the model remove over-dependency on the in-context examples and rather understand the task and answer questions faithfully without being biased by the in-context label distribution. As an ablation, we also remove the task-specific instructions and run the experiments; these experiments are denoted with \textit{wo: without instruction}.

\textbf{Models}: In our study, we utilize a range of LLMs with varying parameter sizes, beginning with
\textit{OpenLlama-7B} \cite{openlm2023openllama}, \textit{MPT-7B} \cite{MosaicML2023Introducing} followed by the even more substantial \textit{OpenLlama-13B} \cite{openlm2023openllama}, \textit{MPT-30B} \cite{MosaicML2023Introducing} and finally \textit{Falcon-40B} \cite{falcon40b}. This progression allows us to explore the impact of increasing model sizes on majority label 
 bias robustness within our study. 
Before running inference using these models, we first instruct-tune them using the OASST \cite{kopf2023openassistant} dataset transforming these base models into instruction-following models.
We limit to opensource models and evaluate model checkpoints from HuggingFace\footnote{\url{https://huggingface.co/}} to enable reproducibility and ensure transparency. 


\begin{figure}[ht]
  \centering
  \includegraphics[width=.48\textwidth]{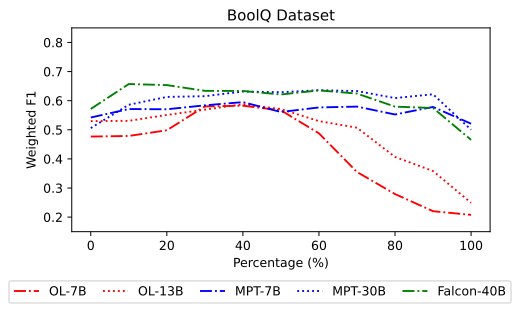}
  \includegraphics[width=.48\textwidth]{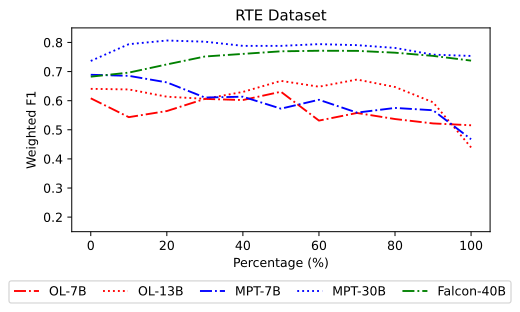}
  \caption{\small [Best viewed in color] \textbf{Comparing various LLMs performance on BoolQ and RTE datasets} - The \% in the X-axis here refers to the \% of \texttt{True} samples and \% of \texttt{Yes} samples for BoolQ and RTE datasets respectively. Here \textit{OL: OpenLlama} models. }
  \label{fig:main}
\end{figure}

\section{Results \& Ablation Study}
\label{results}

\begin{figure*}[ht]
     \begin{minipage}[b]{0.49\textwidth}
        \tiny
\centering
\begin{tabular}
{|p{0.65cm}|p{0.35cm}|p{0.35cm}|p{0.35cm}|p{0.35cm}|p{0.35cm}|p{0.35cm}|p{0.35cm}|p{0.35cm}|p{0.35cm}|p{0.35cm}|p{0.35cm}}
\hline
 & OL-7B-wo & OL-7B-w & OL-13B-wo & OL-13B-w & Falcon-40B-wo & Falcon-40B-w &  MPT-7B-wo & MPT-7B-w& MPT-30B-wo & MPT-30B-w \\ \hline
\multicolumn{11}{|c|}{\textbf{BoolQ}}     \\  \hline
\textit{RB@10 }         & 0.546                & 0.636               & 0.546                 & 0.727                & 0.546                     & \textbf{0.909}                    & 0.455                 & 0.818                & 0.636                  & 0.818                            \\
wt-F1       & 0.450                & 0.443               & 0.469                 & 0.490                & 0.593                     & \textbf{0.605}                    & 0.436                 & 0.548                & 0.545                  & 0.598                               \\
std dev & 0.144                & 0.141               & 0.124                 & 0.107                & 0.112                     & 0.055          &   0.143       & 0.053                 & 0.127                & \textbf{0.049}                                                 \\ \hline
\multicolumn{11}{|c|}{\textbf{RTE}}     \\  \hline
\textit{RB@10 }         & 0.727                & 0.818               & 0.818                 & 0.909                & 0.727                     & \textbf{1.000}                    & 0.546                 & 0.636                & 0.909                  & \textbf{1.000}                             \\
wt-F1       & 0.490                & 0.565               & 0.659                 & 0.618                & 0.706                     & 0.744                    & 0.570                 & 0.601                & 0.756                  & \textbf{0.781}                           \\
std dev & 0.076                & 0.040               & 0.084                 & 0.064                & 0.083                     & 0.031                    & 0.142                 & 0.064                & 0.083                  & \textbf{0.022}                                \\ \hline
\multicolumn{11}{|c|}{\textbf{Covid-5G}}     \\  \hline
\textit{RB@10}          & 0.400                & 0.467               & 0.200                 & 0.267                & 0.333                     & \textbf{0.533}                    & 0.000                 & 0.400                & 0.200                  & 0.467                          \\
wt-F1       & 0.299                & 0.284               & 0.319                 & 0.454                & 0.484                     & \textbf{0.645}                    & 0.341                 & 0.397                & 0.455                  & 0.581                             \\
std dev& 0.123                & 0.066               & 0.123                 & 0.132                & 0.169                     & \textbf{0.079}                    & 0.121                 & 0.106                & 0.183                  & 0.113                               \\ \hline
\end{tabular}
\captionof{table}{Comparison of RB@10, Average and Std. deviation of weighted F1 across various LLMs for the three datasets. The best results are highlighted in \textbf{bold}. Higher the RB@10, the more robust the peak performance.}
\label{tab:rb_f1}
    \end{minipage}
    \hfill
    \begin{minipage}[b]{0.49\textwidth}
        \centering
         \includegraphics[width=.9\textwidth]{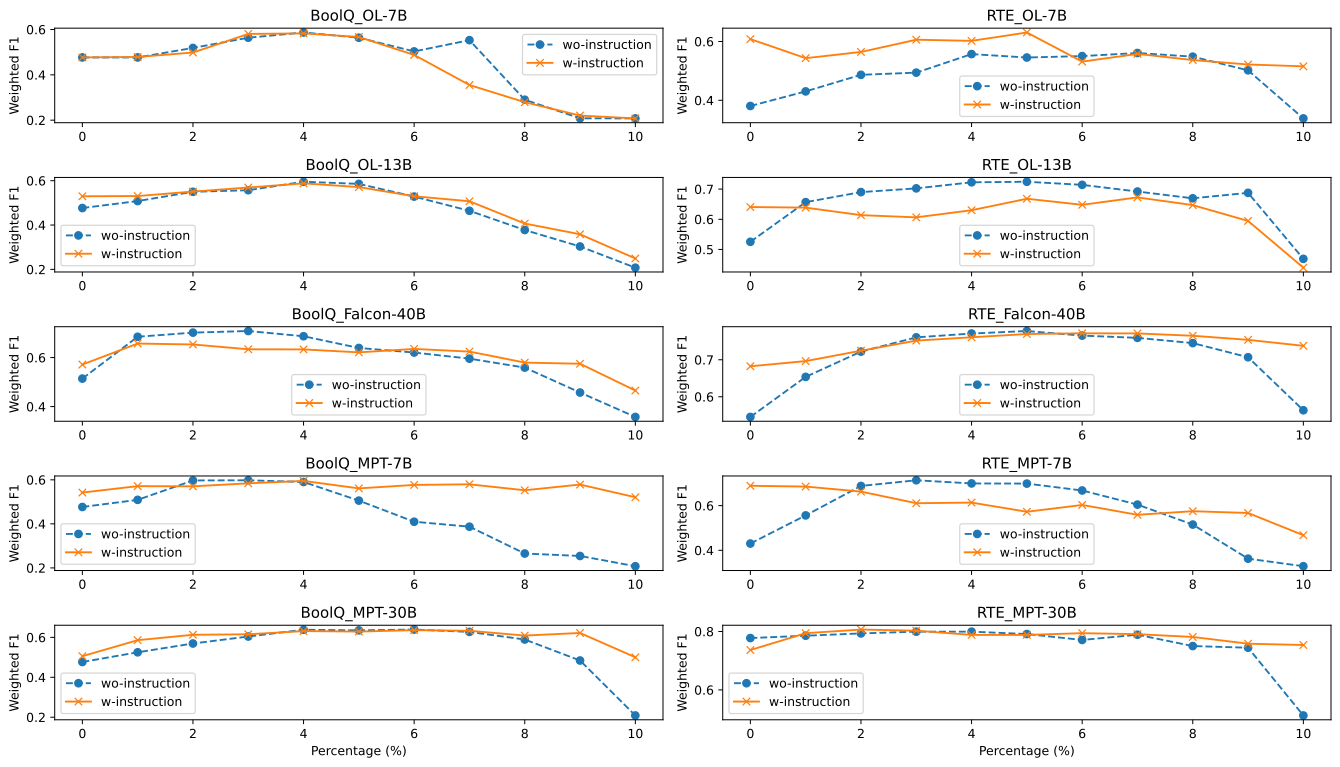}
         \captionof{figure}{Ablating LLM performances \textbf{with} vs \textbf{without instructions}. Here \textit{w: with instruction} and \textit{wo: without instruction}. Also \textit{OL: OpenLlama} models.}
  \label{fig:instruction}
    \end{minipage}
\end{figure*}

The main results are shown in Table~\ref{tab:rb_f1}. All results for BoolQ and RTE datasets are given in Figure \ref{fig:main} and the corresponding ablation study results are given in Figure \ref{fig:instruction}. All results on COVID-5G dataset are given in Table~\ref{tab:table_covid} in the Appendix.

As a means to quantify robustness, we report the $Robustness Boundary \,@ K \,(RB_{K})$ for all models under various majority label distribution settings.
This metric signifies the \% of cases within the sample distribution where the weighted F1-score falls within $\pm K\%$ of the maximum achieved weighted F1-score for the given dataset. The parameter $\#D$ represents the number of distinct distributional settings employed in the prompt. For instance, in datasets such as BoolQ and RTE, $\#D=11$, indicating variations in the proportion of (\texttt{True}, \texttt{False}) across 11 settings, such as $(0\%, 100\%), (10\%, 90\%), \ldots, (100\%, 0\%)$. Similarly, for the Covid-5G Conspiracy dataset, $\#D = 15$.
 Then, $RB_{K}$ is defined as
 
\begin{align}
    \textit{RB}_{K} := \frac{\#(\max_{D}F1 \pm K\% )}{\# D} 
\end{align}


The $RB_{K}$ metric effectively measures the robustness of the peak model performance such that it doesn't deviate more than $K\%$ across different label distribution settings. We choose $K=10$ here and report all results based on that. 
We also report the average and standard deviations results across 5 random seeds for all the models across the three datasets in Table \ref{tab:rb_f1}. 

We report the $RB_{K}$ metric, not just the mean and standard deviation, because in a real-world setting, the critical question from a practitioner's perspective is to what extent a model can perform within an acceptable range given the skewness in the label distribution due to some logistical constraints or inherent data biases. The $RB_{K}$ metric effectively captures this aspect, which is not always evident from the standard deviations of the weighted F1-score.

To illustrate this point, consider the results of \textit{OL-7B w instruction} and \textit{OL-13B w/o instruction} on the BoolQ dataset. The respective mean and standard deviations are (0.443, 0.141) and (0.469, 0.124). However, the $RB_{10}$ metric is 63.6\% and 54.6\%, respectively. As evident, we observe a reversal in the trends of standard deviation and $RB_{10}$. Therefore, even though \textit{OL-13B w/o instruction} has a better F1-score and lower standard deviation compared to \textit{OL-7B w/ instruction}, it exhibits a lower $RB_{10}$ value, indicating potential risks in deploying it in real-world settings. In our study, we aim to highlight these findings, shedding light on the fault tolerance of various LLMs under different distributional settings. One general finding is that \textit{Falcon-40B w/ instruction} and \textit{MPT-30B w/ instruction} consistently performs the best across all three datasets, both in terms of the (mean, std. deviation) of F1-score and the $RB_{10}$ metric.

In a prior work, \citet{zhao2021calibrate} demonstrated that LLMs suffer from majority label bias, meaning they are more inclined to predict the label that is more prevalent in the in-context examples. However, this assertion is only partially true, as we illustrate in Table \ref{tab:rb_f1}, where for binary classification tasks such as BoolQ and RTE, $RB_{10}$ falls within the range of $\sim 50-90\%$, showcasing their robustness to majority label bias. Nevertheless, the bias becomes apparent in scenarios with extreme skewness, resulting in a significant drop in performance (see \textit{OL-7B} in Figure \ref{fig:main}). Notably, for \textit{Falcon-40B w/ instruction}, robustness is maintained even in such extreme cases, with $RB_{10}$ values reaching $\sim 90-100\%$. However, these values decrease significantly to 53.33\% in the case of the COVID-5G dataset, which is a multi-class dataset, indicating reduced robustness. Below we provide some ablations to further illustrate what factors majorly affect model robustness to majority label bias.

    \textbf{Effect of model size and family}: Next, we vary the LLM parameter counts from 7B, 13B, 30B and 40B to study how increasing model size affects the robustness boundary. Here we observe that larger LLMs achieve even higher average weighted F1-score without deviating much from the maximum as we vary the majority label distribution. This can be seen from Figure~\ref{fig:instruction}. For \textit{OpenLlama-13B} the $RB_{10}$ metric is $\sim 1.62\%$ higher compared to \textit{OpenLlama-7B} when averaged across all 3 datasets. This difference further amplifies in case of \textit{MPT-30B} where $RB_{10}$ improves by $\sim 10.51\%$ over the 13B model. The overall best performing model collectively in terms of both classification performance (F1-score) and robustness is \textit{Falcon-40B} where $RB_{10}$ improves $\sim 3.08\%$ over the 30B model averaged across all three datasets. Thus, the trend holds that as we increase the number of model parameters, the robustness metric also increases establishing the fact that larger LLMs are more resilient to majority label bias.
    
    However, the gains are not consistent across different model families. For example, in BoolQ dataset, \textit{MPT-7B w/ instruction} has better robustness compared to \textit{OL-13B w/ instruction}. This can be attributed to the fact that these models are pre-trained with different corpuses and training strategies leading to difference in performance across various benchmarks \cite{naveed2023comprehensive}.

\textbf{w/ instruction vs w/o instruction}: 
We notice that interestingly \textit{w/ instruction} largely outperforms \textit{w/o instruction} variants at the \textbf{tails of the distribution $D$} i.e. extremely skewed cases. This phenomenon may be attributed to the inclusion of task-specific instructions in the prompt, aiding the model in comprehending the task semantics and making predictions rather than over-relying on the in-context prompt examples to understand the task. Compared to \textit{w/ instruction}, the F1-scores for \textit{w/o instruction} variants drops by $\sim 8.3\%$ for all 7B models. The performance gap is further amplified with increasing size of the models. For 13B models the metric drops by $\sim 13.54\%$ and for 30B/40B models it drops by $\sim 27.9\%$. This leads to another finding that larger LLMs are more sensitive to majority label bias in the absence of an informative task-sepcific prompt template. 

Other than the tails, the performance remains mostly similar \textit{w/ or w/o instruction}. On a parallel note, \citet{wei2023larger} show larger LLMs override semantic priors when provided with noisy in-context samples thus deteriorating at the tails. This leads to another finding that larger LLMs are robust to skewed majority label distribution but not to skewed noisy label distribution. This finding is in line with \citet{liu2023good} where they show that the success of in-context learning is more dependent on the input distribution rather than the label distribution.




\section{Conclusion \& Future Work}

In our study, we demonstrate that, contrary to previous findings indicating the lack of robustness of Language Models (LLMs) to majority label bias, there exists a robustness boundary (RB@K) for different LLMs. This boundary is defined as the number of distributionally skewed settings where LLM performance does not deviate from the peak by more than K\% (here K=10). For binary classification tasks, the RB@10 for larger LLMs falls within the range $\sim 80-100\%$, indicating considerable robustness. However, these numbers drop significantly for multi-class classification tasks ($\sim 50\%$), revealing reduced robustness.
We further present ablations that show a positive correlation between this metric and increasing LLM size/parameter count. Additionally, incorporating task-specific instructions in the prompt improves performance in skewed settings.
An intriguing finding is that the impact of adding task-based instructions is more pronounced with larger LLMs compared to smaller ones. This suggests that larger LLMs are more sensitive to instructions given in the prompt.

Given that the final labels are generated in an auto-regressive fashion, our present setup does not provide any control over the generated output. This can be problematic for tasks where the labels are syntactically similar. One way to ensure that the generated text will be always within the set of label names is through Guided Generation \cite{willard2023efficient}, a technique that uses Finite-State-Machine-based vocabulary masking to allow controlled generation from LLMs. Additionally, we are also keen on delving into the robustness of fine-tuning LLMs using PEFT (Parameter-Efficient Finetuning Techniques) \cite{peft} specifically addressing the majority label bias in text classification. 
Beyond the realm of majority label bias, we also intend to broaden our analysis to other types of data \& label biases that can be introduced due to various distribution shifts to provide a comprehensive understanding of the challenges and potential solutions in ensuring model generalization. We leave these discussions for future work.

\section{Acknowledgements}
We would like to express our gratitude to Bing He for generously sharing the dataset used in their paper \citet{micallef2020role}.

\bibliography{aaai24}

\appendix

\begin{table*}[ht]
\tiny
\centering
\begin{tabular}
{p{0.75cm}|p{0.5cm}|p{0.5cm}|p{0.5cm}|p{0.5cm}|p{0.5cm}|p{0.5cm}|p{0.5cm}|p{0.5cm}|p{0.5cm}|p{0.5cm}|p{0.5cm}}
proportions (\%,\%,\%) & OL-7B-wo & OL-7B-w & OL-13B-wo & OL-13B-w & Falcon-40B-wo & Falcon-40B-w & MPT-7B-wo & MPT-7B-w& MPT-30B-wo & MPT-30B-w \\ 
\hline
0,0,100     & 0.157                & \cellcolor{yellow!=45}0.312               & 0.157                 & 0.211                & 0.157                     & 0.473                                  & 0.158                 & 0.316                & 0.169                  & 0.415                                   \\
100,0,0     & 0.161                & 0.193               & 0.161                 & 0.419                & 0.501                     & 0.569                                & 0.161                 & 0.310                & 0.172                  & 0.436                         \\
0,100,0     & 0.184                & 0.264               & 0.248                 & 0.276                & 0.258                     & 0.600                                  & 0.183                 & 0.314                & 0.248                  & \cellcolor{yellow!=45}0.649                                    \\ \hline \hline
0,25,75     & 0.163                & 0.261               & 0.186                 & 0.327                & 0.264                     & 0.635                                & 0.411                 & 0.303                & 0.455                  & 0.493                                   \\
0,75,25     & 0.300                & 0.281               & 0.241                 & 0.348                & 0.450                     & \cellcolor{yellow!=45}0.651                                  & 0.324                 & 0.327                & 0.457                  & \cellcolor{yellow!=45}0.715                                    \\
25,0,75     & \cellcolor{yellow!=45}0.431                & \cellcolor{yellow!=45}0.351               & 0.304                 & 0.461                & 0.340                     & 0.563                                  & 0.297                 & \cellcolor{yellow!=45}0.505                & 0.434                  & 0.440                                   \\
75,0,25     & \cellcolor{yellow!=45}0.390                & 0.200               & 0.335                 & 0.483                & 0.507                     & 0.613                                 & 0.440                 & \cellcolor{yellow!=45}0.510                & 0.454                  & 0.529                                 \\
75,25,0     & 0.175                & 0.177               & 0.372                 & 0.482                & \cellcolor{yellow!=45}0.618                     & \cellcolor{yellow!=45}0.710                                 & 0.269                 & 0.284                & 0.428                  & 0.529                                     \\
25,75,0     & 0.240                & 0.259               & 0.390                 & \cellcolor{yellow!=45}0.549                & 0.519                     & \cellcolor{yellow!=45}0.730                                 & 0.315                 & 0.356                & 0.431                  & \cellcolor{yellow!=45}0.667                                  \\ \hline \hline
25,25,50    & \cellcolor{yellow!=45}0.450                & \cellcolor{green!=45}0.396               & \cellcolor{green!=45}0.535                 & \cellcolor{yellow!=45}0.616                & \cellcolor{yellow!=45}0.652                     & \cellcolor{yellow!=45}0.721                                 & 0.453                 & \cellcolor{yellow!=45}0.518                & \cellcolor{yellow!=45}0.720                  & \cellcolor{yellow!=45}0.695                                    \\
25,50,25    & \cellcolor{green!=45}0.470                & \cellcolor{yellow!=45}0.296               & \cellcolor{yellow!=45}0.527                 & \cellcolor{yellow!=45}0.636                & \cellcolor{yellow!=45}0.710                     & \cellcolor{green!=45}0.742                                  & 0.412                 & \cellcolor{yellow!=45}0.535                & \cellcolor{yellow!=45}0.763                  & \cellcolor{green!=45}0.746                                   \\
50,25,25    & \cellcolor{yellow!=45}0.463                & \cellcolor{yellow!=45}0.306               & \cellcolor{yellow!=45}0.435                 & \cellcolor{green!=45}0.643                & \cellcolor{green!=45}0.715                     & \cellcolor{yellow!=45}0.729                                 & \cellcolor{green!=45}0.589                 & \cellcolor{green!=45}0.548                & \cellcolor{green!=45}0.764                  & \cellcolor{yellow!=45}0.685                                   \\ \hline \hline
50,0,50     & \cellcolor{yellow!=45}0.402                & \cellcolor{yellow!=45}0.335               & 0.306                 & 0.483                & 0.536                     & \cellcolor{yellow!=45}0.642                              & 0.439                 & \cellcolor{yellow!=45}0.509                & 0.448                  & 0.475                                 \\
0,50,50     & 0.227                & \cellcolor{yellow!=45}0.384               & 0.193                 & 0.342                & 0.427                     & 0.587                                & 0.375                 & 0.295                & 0.431                  & 0.596                                   \\

50,50,0     & 0.273                & 0.252               & 0.392                 & 0.533                & \cellcolor{yellow!=45}0.616                     & \cellcolor{yellow!=45}0.715                              & 0.293                 & 0.329                & 0.444                  & \cellcolor{yellow!=45}0.651                                   \\

\end{tabular}
\caption{ \small Comparing various LLMs performance on \textbf{Covid-5G dataset} for the varying label proportions \textit{(misinformation, counter-misinformation, irrelevant)\%} in the prompt. Here \textit{w: with instruction} and \textit{wo: without instruction}. Also \textit{OL: OpenLlama} models. The optimal performance levels for each model, highlighted in \textcolor{green}{green}, tend to be distributed predominantly around the central region where there is a discernible presence of each label. Additionally, points within the RB@10 boundary, depicted in \textcolor{yellow}{yellow}, are scattered across the spectrum. }
\label{tab:table_covid}
\end{table*}
\newpage
\section{Appendix: Implementation Details}
\label{sec:implement}

For the BoolQ dataset, we conducted a series of experiments with various LLMs as mentioned in Section \ref{proposed}. In these experiments, we implemented two distinct prompt strategies: (a) one that incorporated task-specific instructions + in-context examples (b) another that presented only in-context examples without instructions. In (a), the prompt was: \textit{``You need to classify the following sentence as True or False. Below, some examples are provided"}. Subsequently, we appended examples along with their corresponding ground truth in the format: \textit{``Sentence:"} + [Sentence] + \textit{``output:"} + [Answer]. We provide 50 such example sentences in the prompt from the training data. At the end of the prompt, we append a sentence from the test dataset in a similar format but without providing the true label. The LLM is expected to generate the label based on the instruction and in-context examples. In the LLM inference pipeline, we set $max\_new\_tokens = 2$. We found that almost always the generated text belonged to the set of labels and in only $< 3\%$ of test datapoints the LLMs hallucinate. In the alternative setting, devoid of instructions, we included examples in the prompt, such as \textit{``Sentence:"} + [Sentence] + \textit{``output:"} + [Answer]. 

Similarly, when working with the RTE dataset, we adopted a configuration analogous to that employed for the BoolQ dataset. We conducted comparable experiments with two prompt strategies: (a) incorporating explicit instructions and examples and (b) presenting examples alone. The initial prompt included the instruction: \textit{``You need to classify the following premise-hypothesis pair as True or False. Below, some examples are provided"}. Subsequently, examples were included along with their corresponding ground truth in the format: \textit{``Premise"} + [PRE] + \textit{``hypothesis"} + [HYP] + \textit{``entailment"} + [ENT]. In the alternative setting, without instructions, examples were included in the prompt as follows: \textit{``Premise"} + [PRE] + \textit{``hypothesis"} + [HYP] + \textit{``entailment"} + [ENT]. The LLM inference setup is similar to the one described above for BoolQ dataset. 

\black{
The setting for Covid 5G misinformation dataset is analogous to BoolQ and RTE dataset. We conducted experiments with two prompt strategies: (a) incorporating explicit instructions and examples as follows - ``You need to classify a following tweet into one of the following - misinformation, correct or irrelevant. Below some examples are provided.''; subsequently example tweets are included in the following format  \textit{``Tweet : "} + [tweet] + \textit{``Label:"} + [HYP] . (b) without incorporating explicit instructions but with just examples in the above format. The rest of LLM inference pipeline is similar to the other two datasets. 
}

\end{document}